\documentclass{article}

    \usepackage[preprint,nonatbib]{neurips_2020}

\usepackage[utf8]{inputenc} 
\usepackage[T1]{fontenc}    
\usepackage{hyperref}       
\usepackage{url}            
\usepackage{booktabs}       
\usepackage{amsfonts}       
\usepackage{nicefrac}       
\usepackage{microtype}      

\usepackage{amsmath}
\usepackage{amssymb}
\usepackage[ruled, linesnumbered, vlined]{algorithm2e}
\usepackage{graphicx}
\usepackage{multirow}
\usepackage{subfigure}
\usepackage{multicol}
\usepackage{threeparttable}
\usepackage{booktabs}
\usepackage{color}
\usepackage{authblk}
\usepackage{amsthm}

\theoremstyle{definition}
\newtheorem{definition}{Definition}[section]

\newcommand{\F}{\textsl{FMA-ETA}\xspace}

\newcommand*{\affaddr}[1]{#1} 
\newcommand*{\affmark}[1][*]{\textsuperscript{#1}}
\newcommand*{\email}[1]{\texttt{#1}}

\title{FMA-ETA: Estimating Travel Time Entirely \\ Based on FFN With Attention}

\author{%
\textbf{Yiwen Sun\affmark[1] \  Yulu Wang\affmark[1] \  Kun Fu\affmark[2] \  Zheng Wang\affmark[2] \  Ziang Yan\affmark[1]}\\ \textbf{Changshui Zhang\affmark[1] \  Jieping Ye\affmark[2]}\\
\affaddr{\affmark[1]Department of Automation, Institute for Artificial Intelligence\\ Tsinghua University (THUAI), Beijing, China}\\
\affaddr{\affmark[2]DiDi AI Labs, Beijing, China}\\
\email{syw17@mails.tsinghua.edu.cn, wangyulu18@mails.tsinghua.edu.cn}\\ 
\email{fukunkunfu@didiglobal.com, wangzhengzwang@didiglobal.com}\\
\email{yza18@mails.tsinghua.edu.cn, zcs@mail.tsinghua.edu.cn, yejieping@didiglobal.com}\\
}
\begin{document}

\maketitle

\begin{abstract}
  Estimated time of arrival (ETA) is one of the most important services in intelligent transportation systems and becomes a challenging spatial-temporal (ST) data mining task in recent years. Nowadays, deep learning based methods, specifically recurrent neural networks (RNN) based ones are adapted to model the ST patterns from massive data for ETA and become the state-of-the-art. However, RNN is suffering from slow training and inference speed, as its structure is unfriendly to parallel computing. To solve this problem, we propose a novel, brief and effective framework mainly based on feed-forward network (FFN) for ETA, FFN with Multi-factor self-Attention (\F). The novel Multi-factor self-attention mechanism is proposed to deal with different category features and aggregate the information purposefully. Extensive experimental results on the real-world vehicle travel dataset show \F is competitive with state-of-the-art methods in terms of the prediction accuracy with significantly better inference speed.
\end{abstract}

\section{Introduction}
\label{sec:Introduction}
Estimated time of arrival (ETA) or travel time prediction is universally considered as the travel time estimation given a pair of origin and destination locations along the route~\cite{wang2018learning}. As an essential component of artificial intelligence for transportation, ETA influences route planning, navigation and vehicle dispatching which are fundamental for ride-hailing platforms, such as DiDi and Uber~\cite{wang2018learning,wang2018will}.
ETA is a representative and challenging sequence learning and data mining task attracting lots of attention~\cite{wang2014travel,hofleitner2012learning,chen2013dynamic,zhang2016urban,wang2018learning,wang2018will,li2018multi}. 

Since 2018, deep learning~\cite{lecun2015deep} based methods~\cite{wang2018will,li2018multi,wang2018learning} which significantly overperform non-deep learning-based methods~\cite{wang2014travel,hofleitner2012learning,chen2013dynamic,zhang2016urban} mines the spatial-temporal correlations concurrently and effectively from large-scale data and become state-of-the-art.
The general sequential semantic information extractor of these state-of-the-art methods, such as WDR~\cite{wang2018learning}, DeepTTE~\cite{wang2018will}, DeepTravel~\cite{zhang2018deeptravel} are mainly one Recurrent Neural Network (RNN)~\cite{hopfield1982neural,jordan1997serial,elman1990finding} variant, Long Short-Term Memory Network (LSTM)~\cite{hochreiter1997lstm}. RNN adopts the recurrent structure to model sequence and extract semantic information, which also determines its restricted inference speed due to non-parallelization.

\begin{figure}[htb]
    \centering
\includegraphics[width=0.9\linewidth]{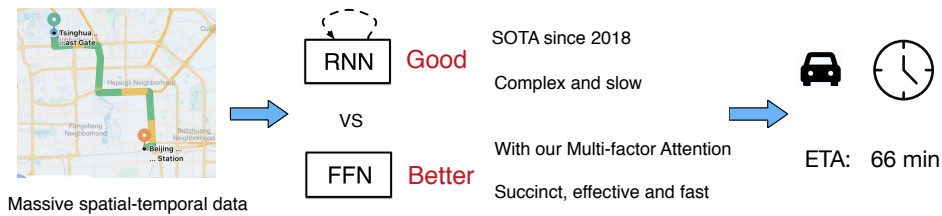}
\caption{The conceptual demonstration of ETA and two kinds of candidate sequence feature extractors based on deep learning. ETA refers to estimating the travel time along the given route between the origin and destination. In real application scenarios, the RNN, such as current state-of-the-art LSTM is copmlex and slow. FFN with our proposed Multi-factor Attention (\F) is promising for the future of ETA due to simplicity, high speed and effectiveness.}
\label{fig:sketch}
\end{figure}

In this paper, we discuss the possibility of mainly adopting FFN to mine spatial-temporal  information from sequential massive data for ETA, as illustrated in Fig.~\ref{fig:sketch}. FFN is parallelizable and naturally beneficial for fast ETA inference considering accuracy simultaneously which is a industry pain point for ride-hailing platforms. However, completely depending on FFN, the model can hardly capture the dependency between links.

Will there be a novel structure helping FFN analyze sequence semantic information effectively and play its obvious advantages in inference speed for ETA? 
Follow this line, we present a novel Multi-factor Attention which is specially designed for ETA, a sequential learning task affected by various factors.
\F which is mainly based on FFN with Multi-factor Attention is proposed for a better sequence feature extractor than RNN which is state-of-the-art since 2018. 

The main contributions in this work are as follows:

\begin{itemize}
    
    \item We propose a novel ETA deep learning based framework, \F that is the first deep learing framework entirely based on FFN with attention, to our best knowledge.

    \item We propose a novel Multi-factor Attention mechanism for effectively learning the time dependency and semantic information between time steps of the sequence. Through sufficient experiments, we find that for ETA, Multi-factor Attention is better than Multi-head attention~\cite{vaswani2017attention} which is famous in natural language processing. Besides, 
    Multi-factor attention can be adpoted and may also be promising for other sequence learning tasks affected by various factors.

    \item We evaluate \F on the massive real-world dataset containing over 500 million trajectories from one famous ride-sharing platform. 
    The abundant experimental results demonstrate that \F's estimation precision  is comparable with the state-of-the-art RNN based method, WDR.
    Not only that, \F improves the ETA model inference speed than WDR significantly.
    
\end{itemize}

We organize the paper as follows. Section~\ref{Background} briefly summarizes the backgrounds of ETA, sequence learning and attention mechanism. Section~\ref{Model} introduces the overall framework of \F, followed by the description of the general Multi-factor attention in detail.
In Section~\ref{Why}, we elaborate the reason why we propose Multi-factor attention.
In Section~\ref{exp}, experimental result comparisons on the large-scale real-world dataset are presented to show the excellent accuracy and inference speed of \F. Finally, this paper is concluded and the possibility of further work is analyzed in Section~\ref{Conclusion}.

\section{Background}
\label{Background}
In this section, we briefly overview the background of our work, inculding estimated time of arrival and attention mechanism.
\subsection{Estimated time of arrival}
Estimated time of arrival (ETA) is a challenging problem in the field of intelligent transportation system.
There are two representative methods for solving ETA, route-based method and data-driven method. The route-based method focus on formulate the travel time of a given route as the summation of time on each road segment and each intersection. Traditional machine learning methods such as dynamic Bayesian network \cite{hofleitner2012learning},
least-square minimization \cite{zhan2013urban} and pattern matching \cite{chen2013dynamic} are typical approaches to capture the spatial-temporal features in the route-based method. However, the idea of dividing the original trajectory results in the accumulation of local errors. The data-driven method has shifted from traditional methods such as TEMP \cite{wang2019simple} and time-dependent landmark graph \cite{yuan2011t} to deep-learning based methods \cite{wang2018learning,fu2020}. MURAT \cite{li2018multi} uses multi-task learning and graph convolutional networks to assist a residual block to predict the travel time from the departure to the destination without a given trajectory. 
In recent years, researchers have conducted more explorations on applying deep learning methods to solve ETA problems, such as Deeptravel \cite{zhang2018deeptravel}, DeepTTE \cite{wang2018will}, Deepi2t \cite{lan2019travel} and WDR \cite{wang2018learning}. These methods apply different approaches on modeling spatial information, but they all use LSTM \cite{hochreiter1997lstm} to extract features from time series. However, the inference speed of the model with LSTM is too slow to be applied in actual scenarios. 
In this work, we proposed \F which can sufficiently handle the above problem.

\subsection{Attention mechanism}
Attention is a very effective mechanism in natural language processing \cite{bahdanau2014neural}, image caption \cite{xu2015show} and other research areas \cite{ren2019fastspeech}.  Attention mechanism has outstanding ability in capturing semantic dependencies. 
Common attention mechanisms are local attention, global attention \cite{luong2015effective}, self-attention \cite{vaswani2017attention}, etc. 
Transformer \cite{vaswani2017attention} is a novel sequence to sequence network entirely based on FFN with Multi-head self-attention. It achieves promising results in translation with a faster speed than RNN-based models. Then self-attention becomes a hot topic in neural network attention research. Self-attention is calculated by:
\begin{equation}
\text { Self-attention }=\operatorname{softmax}\left(\frac{\mathbf{Q} \mathbf{K}^{T}}{\sqrt{d}}\right) \mathbf{V}
\end{equation}
where $\mathbf{Q},\mathbf{K},\mathbf{V}$ is query, key and value matrix, $d$ is the dimension of key and query matrix, and key and query matrix are usually the same.  Self-attention is proved useful in a wide variety of tasks including sequential recommendation \cite{kang2018self}, reading comprehension \cite{yu2018qanet}, speech recognition \cite{salazar2019self} and traffic flow predicting \cite{zhu2018end}.

Deep learning-based ETA models are mostly RNN-based models. RNN has problems when deals with long-range dependencies. LSTM are able to deal the problem to some extent, but in practice it still have problems in long-range dependencies. The inference time of LSTM-based model is too long for practical application, so it is vital to introduce attention mechanism to the ETA problem.

\section{Model Architecture}
\label{Model}
We first give the accurate mathematical definition of estimated time of arrival with reference to \cite{wang2018learning}.
\begin{definition}[\textbf{Estimated time of arrival}]\label{defETA}
For a collection of historical trips $D=\left\{s_{i}, e_{i}, d_{i}, \boldsymbol{p}_{i}\right\}_{i=1}^{N}$, where $s_{i}$ is the departure time for the i-th trajectory, $e_{i}$ is the arrival time for i-th trajectory, $d_i$ is the driver ID, $\boldsymbol{p}_{i}$ is the link sequence set of the trajectory, and $N$ is the total number of samples. The ground truth travel time is computed by $y_{i}=e_{i}-s_{i}$. Here the link sequence set $\boldsymbol{p}_{i}$ can be represented as $\boldsymbol{p}_{i}=\left\{l_{i 1}, l_{i 2}, \cdots, l_{i T_{i}}\right\}$, where $l_{ij}$ represents the j-th link in the i-th trajectory, and $T_i$ is the length of the link sequence set.
\end{definition}
Sicne 2018, most state-of-the-art methods' main force for capturing spatial-temporal patterns to complete ETA has change into RNN (specifically, LSTM). RNN is a famous general sequence feature extractor for various sequence learning subfields, such as speech signal processing and natural language processing.
In this paper, we break the stereotype and present a ETA framework entirely based on FFN and novel Multi-factor Attention, \F.  We introduces the  overall structure of \F as well as proposed Multi-factor attention in next two subsections. 

\subsection{Overall framework}
The first main step is the sophisticated feature engineering where we follow~\cite{wang2018learning}.
Rich features from massive raw data is the key input for deep learning model.
Features could be divided into the following two categories.

(1) Global features are sparse and one trajectory corresponds to one global representation, such as driverID, day of week, departure time slice.
The method, Embedding~\cite{bengio2003neural} is adopted for the dimensionality reduction of sparse features. 

(2) Sequential features are related to each link of the trajectory, for instance, length of the link (road segment), speed (road contidion), link time (related to road contidion) and embedding of linkID. These four factors influence ETA from different perspectives.

\begin{figure}[htb]
    \centering
\includegraphics[width=0.98\linewidth]{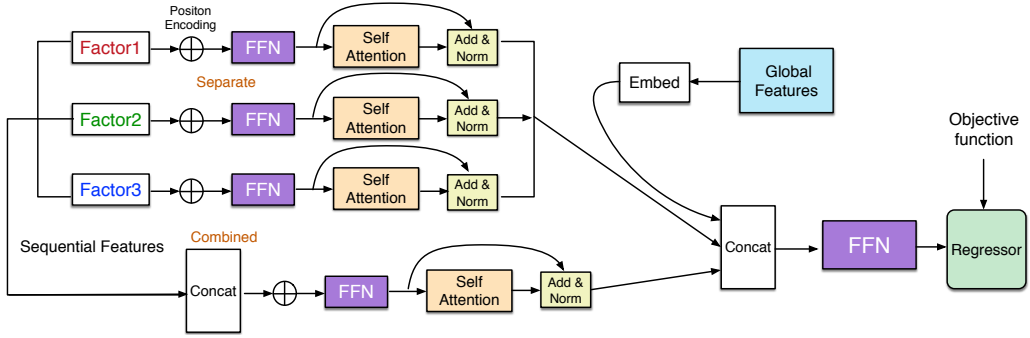}
\caption{\F two main components for sequential features: FFN and Multi-factor Attention.}
\label{fig:FMA-ETA}
\end{figure}

We then describe the overall framework of \F, as shown in Fig.~\ref{fig:FMA-ETA}.
Two main components of \F are Multi-factor Attention and FFN.

(1) Sequential features adopt our Multi-factor Attention which will be discussed in detail in next subsection. This component fully explores the relationship between different links in each track.

(2) Parallelizable FFN is the main reason for simplicity and fast inference that are our greatest advantages compared with RNN.
The front FFN is utilized for each sequential factor to mining the spatial-temporal patterns in their single aspect as well as for concatenated factors.
The last FFN is for the information aggregation from sequential separate and combined representations as well as embeddings of global features.

The regressor is one linear layer with ReLU~\cite{krizhevsky2012imagenet} as the activation function.
The Objective function of the overall deep learning model is the mean absolute percentage error (MAPE) which is an common and relative loss function for ETA. The \F's parameters are trained through:

\begin{equation}
\min _{\theta} \sum_{i}\left|\frac{y_{i}-y_{i}^{\prime}}{y_{i}}\right|,
\end{equation}
where $ y_{i}^{\prime}$ is $i_{th}$ query's ETA, $y_i$ is the ground truth time and $\theta$ is all the parameters of \F.

\subsection{Multi-factor Attention}
ETA is a challenging and complex problem due to the fact that various factors affect the accuracy of prediction, such as the link length as well as its road condition. Therefore, unlike for natural language processing where one word can be represented by a single embedding vector, different sequential features ought to be treated and dealt with more specifically for ETA.
Our Multi-factor Attention mechanism is proposed to let different sequential factors mine its patterns and the impact on ETA in different subspaces, as shown in the upper left corner of Fig.~\ref{fig:FMA-ETA}. Self attention is with reference to~\cite{vaswani2017attention} and we add position encoding, residual connections, layer normalization, and dropout after self attention following~\cite{vaswani2017attention}.
Combined sequential features also capture the spatial-temporal patterns as a whole by FFN with self-attention. 

In the Fig.~\ref{fig:FMA-ETA}, we show the Multi-factor Attention mechanism with three factors. When the number of the factors is arbitrary $n$, the general Multi-factor Attention could be expressed:
\begin{equation}
\begin{aligned}
\text {Multi-factor}(\text{factor}_1, ..., \text{factor}_n) \left.=\text{Concat}(\text{self}_1, ..., \text{self}_n, \text{self}_\text{all}\right) \mathbf{W}^{O} \\
\text {where} \ \text{self}_i =\text {Self-attention}\left(\text{factor}_i \mathbf{W}_{i}^{Q}, \text{factor}_i \mathbf{W}_{i}^{K}, \text{factor}_i \mathbf{W}_{i}^{V} \right),\\
 i=1,...,n\\
\text{self}_\text{all} =\text {Self-attention}\left(\text{factor}_C \mathbf{W}_{all}^{Q}, \text{factor}_C \mathbf{W}_{all}^{K}, \text{factor}_C \mathbf{W}_{all}^{V} \right), \\
\text{factor}_C = \text{Concat}(\text{factor}_1, ..., \text{factor}_n).
\end{aligned}
\end{equation}
Where $\text{factor}_i$ is the i-th factor of ETA problem, ${\mathbf{W}_i}^{*}$ is the learned parameters in the FFN layers of the i-th factor, ${\mathbf{W}_{all}}^{*}$ is the learned parameters in the FFN layers of the combined features.
For our \F, sequential factors contains (1) length of the link, (2) road contidion speed, (3) corresponding link time and (4) embedding of linkID. Hence, we adopt the Multi-factor Attention version of four factors, i.e., $n = 4$.
Through concatenating the separate and combined sequence representations, our Multi-factor Attention complete the multi level and detailed extraction of the spatial-temporal dependencies of sequence data.

\section{Why Multi-factor Attention}
\label{Why}
In this section, we will discuss the motivation and reason for the proposal of multi-factor attention. The RNN-based model has a good performance on the ETA problem of which the evaluation metric is good. However, the RNN-based model has a slow training/inference speed, making it difficult to be applied in practical problems. FFN is a promising method to accelerate the speed of the model. But FFN has a poor performance in sequence learning and have problems on long-range dependencies. Our proposed multi-factor attention can solve the above problem. We will analyze and compare the total computational complexity and sequential operations of RNN and FFN with Multi-factor Attention.

As shown in Table \ref{tab_comp}, the multi-factor attention only need $O(1)$ sequetial operations while RNN requires $O(n)$. As for computational complexity, when the length of sequence $n$ is smaller than the dimension of features $d$, our multi-factor attention is faster than RNN. 

Self attention especially multi-head attention has achieved good results on sequence learning. Why not multi-head attention? In terms of ETA problem, there are many different factors affect it and the traffic state is complex and dynamic. Experiments in Section \ref{exp} show that multi-head attention does not perform well on complex problems in the transportation system like ETA. Our multi-factor attention focus on both separate features and combined features. In this way can we promote different subspaces to analyze the effect of a certain factor pattern on ETA. The evaluation metrics shows that model with multi-factor attention preforms better than model with multi-head attention on ETA problems. 

Hence, multi-factor attention is more effective for extracting systematic and comprehensive spatial-temporal patterns comparing with multi-head attention. Considering the speed promotion of FFN, FFN with multi-factor attention has a great advantage in tasks in intelligent transportation system (ITS). Multi-factor attention is a general method and may be also promising for other time series forecasting tasks.
\begin{table}[!t]
\centering
\caption {The per layer complexity and sequential operations of different methods}
\label{tab_comp}
\footnotesize
\begin{threeparttable}
\begin{tabular*}{0.73\textwidth}{ccc}
\toprule
               & Complexity per Layer &  Sequential Operations\\
\midrule
 RNN   & $\mathcal{O}(n^2d)$  &   $\mathcal{O}(n)$ \\
 Multi-factor Attention       & $\mathcal{O}(nd^2)$ &   $\mathcal{O}(1)$ \\
\bottomrule
\end{tabular*}
\begin{tablenotes}
    \footnotesize
     \item[*] $n$ is the length of the sequence, $d$ is the dimension of features.
\end{tablenotes}
\end{threeparttable}
\end{table}

\section{Results}
\label{exp}
\subsection{Dataset}
We evaluate our model on a large-scale real-world floating-car trajectory dataset Beijing 2018 collected by a ride hailing platform. It contains the trajectory data of hundreds of millions of Beijing taxi drivers after desensitization for more than 4 months in 2018. This dataset covers different types of roads in Beijing urban areas, including local streets and freeways. We filter out the abnormal data where the driving time is less than 60s or the speed exceeds 120 km/h in Beijing 2018. We divided this data set into atraining set (the first 16 weeks of data), a validation set (the middle2 weeks of data) Test set (data for the last 2 weeks).
\subsection{Compared methods}
We compared the proposed \F with the following competitors:

(1) \textbf{Route-ETA}: a representative method for traditional non-deep learning methods. It divides the trajectory into several links and intersections. The travel time $t_i$ in the i-th link of this trajectory is calculated by dividing the link's length by the estimated speed in the i-th link. The delay time $c_j$ in the j-th intersection is provided by a real-time traffic monitoring system.The final arrival time is the sum of the estimated time spent in each subsection.

(2) \textbf{WDR(RNN)}: a deep learning method achieving the state-of-the-art performance in ETA problem. WDR is a joint model including width module, depth module, and recurrent module. It can effectively use the dense features, high-dimensional sparse features and local features of road sequence in traffic information. Here we use RNN in the recurrent module.

(3) \textbf{WDR(LSTM)}: a variants of WDR(RNN).  Here we use LSTM in the recurrent module of WDR and we make no changes to other part of WDR. 

(4) \textbf{WD-FFN}: a variants of WDR. It uses deep module to replace the recurrent module. Here we use a Multi-Layer Perceptron network for comparision.

(5) \textbf{WD-Resnet}: a variants of WDR. It uses deep module to replace the recurrent module. Here we use a residual structure to extract features.

(6) \textbf{Multi-head attention}: a variants of WDR. We use FFN with multi-head attention mechanism instead of RNN to extract features from the sequential data.

If this work is accepted, we will open source the codes of proposed deep learning-based model, \F.
\subsection{Experimental Settings}
In our experiment, all models are written in PyTorch. They are trained and evaluated on a single NVIDIA Tesla P40 GPU. The number of iterations of the deep learning-based method is 3.5 million. We use the method of Back Propagation (BP) to train the deep learning-based methods, and the loss function is the MAPE loss. We choose Adam as the optimizer due to its good performance. The batch size is 256 and the initial learning rate is 0.0002.

\subsection{Evaluation Metrics}
To evaluate and compare the performance of \F and other baselines, we use evaluation metrics, Mean Absolute Percentage Error (MAPE), Mean Absolute Error (MAE) and Rooted Mean Square Error (RMSE):
\begin{equation}
\mathrm{MAE}(y,y^{\prime})=\frac{1}{N} \sum_{i=1}^{N}\left|y_{i}-y_{i}^{\prime}\right|
\end{equation}
\begin{equation}
\mathrm{RMSE}(y,y^{\prime})=\sqrt{\frac{1}{N} \sum_{i=1}^{N}\left(y_{i}-y_{i}^{\prime}\right)^{2}}
\end{equation}
where $ y_{i}^{\prime}$ is the predicted travel time, and $y_i$ is the ground truth travel time.
The calculation process of MAPE is shown in Section 3.

\subsection{Experimental Results and Analysis}

\begin{table}[!t]
\centering
\caption {The results of different methods}
\label{tab_loss}
\footnotesize
\begin{threeparttable}
\begin{tabular*}{0.73\textwidth}{ccccc}
\toprule
               & MAE(sec) &  RMSE(sec) & MAPE(\%)& Latency(ms) \tnote{*}\\
\midrule
 Route-ETA  & 69.008 & 106.966 & 25.010& \textbf{0.179}\\
 WD-FFN      &  57.797    &   93.588&   21.106& 0.344\\
 WD-Resnet   &  57.064 &     92.241 &   21.015      & 0.454\\
 WDR(RNN)  &55.284 & 90.836 & 19.677& 1.107\\
 WDR(LSTM)    & 55.227  &   90.480 &   \textbf{19.598}     & 1.109\\
 Multi-head attention       & 55.145  &   90.101  &   19.678 & 0.635\\
 \textbf{FMA-ETA (ours)}    &  \textbf{54.642}&     \textbf{88.794}   &  19.618       & 0.866\\
\bottomrule
\end{tabular*}
\begin{tablenotes}
    \footnotesize
     \item[*] Latency is the average inference time of the models.
\end{tablenotes}
\end{threeparttable}
\end{table}
Table \ref{tab_loss} shows the general three evaluation metrics for ETA problems. Our \F outperforms all competitors in terms of MAE and RMSE metrics. \F achieves similar results with the start-of-the-art method WDR(LSTM) in terms of MAPE metric. The detailed analysis of the experimental results are as follows.

 (1) The representative non-deep learning method, route-ETA performs worse than other deep learning based methods. It shows that the data-driven method is more effective than route-based method. The deep method is suitable for modeling complex transportation system given massive spatio-temporal data.

 (2) Models with recurrent module performs better than models that only use deep modules without attention mechanism. WDR(RNN) and WDR(LSTM) achieves better results than WD-FFN and WD-Resnet. WDR(LSTM) performs best on MAPE metric, because the use of gated units can solve the problem of long-term dependencies to a certain extent. The deep modules with attention achieve better results than WDR on MAE and RMSE metrics, which means attention mechanism can help to extract features and sole the long-range dependencies in long sequence.

 (3) Our \F performs best on MAE and RMSE metrics, which means our method is very applicable to ETA problems. Our FMA-ETA outperforms LSTM by 1.05\% in terms of MAE loss and 1.86\% in terms of RMSE loss. Our \F perform similar results to WDR(LSTM), and our \F only 0.1\% worse than WDR(LSTM) in terms of MAPE metric. Considering the three evaluation metrics, our \F performs best on ETA problems.

 (4) As can be seen in the "Latency" column of Table \ref{tab_loss}, our \F speed up the inference process by $21.8\%$ compared with WDR(LSTM). Route-ETA has the shortest time of 0.179s, but its performance on evaluation metrics is poor. FFN-based methods without attention mechanism is fast, but it brings a great loss on the evaluation metrics. Model with multi-head attention is faster than \F. Its performance is worse than \F. If the performance of models for ETA problem is not good enough, many tasks of ETA's downstream in intelligent transportation systems such as route planning, navigation and vehicle dispatching will be affected greatly. Therefore, we should increase the inference speed of the model while ensuring the accuracy. Currently only our \F can reach the goal.

 Our \F has a good performance on the ETA problem, and it greatly prompts the inference speed compared with the state-of-the-art method WDR(LSTM). \F achieves clear improvements over WDR(LSTM) regarding to MAE and MAPE metrics. Taking into account both evaluation metrics and speed, our method is the most suitable method for ETA problems.
 
\subsection{Speed comparison of different methods}
\begin{figure}[t]
\includegraphics[width=0.8\linewidth]{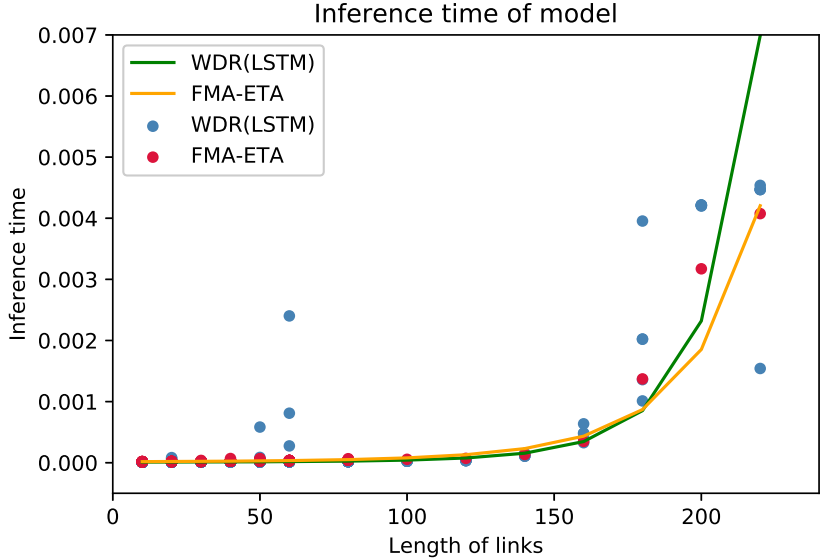}
\caption{The inference time of WDR(LSTM) and FMA-ETA.}
\label{fig:inference}
\end{figure}
As we analyzed above, the state-of-the-art method WDR(LSTM) for ETA problem in previous literature takes a long time for inference. This makes WDR(LSTM) hard to be applied in the real-time traffic system. FFN can greatly accelerate the inference speed of the model for ETA problem, such as WD-FFN and WD-Resnet, but it causes a large decrease in accuracy which can be seen in Table \ref{tab_loss}. The attention mechanism can help FFN to effectively extract sequence features. The existing multi-head attention improves the inference speed, but it still brings a great loss of accuracy. Our goal is to increase the inference speed of the ETA model while ensure that the evaluation metrics do not decrease. We can see from the average inference speed in Tabel \ref{tab_loss}, only \F can achieve the goal. We further explore the inference speed of WDR(LSTM) and \F with different sequence length. We randomly sample 50 samples at each sequence length for WDR(LSTM) and \F, then plot the scatters in Figure \ref{fig:inference}. The curve in Figure \ref{fig:inference} is obtained by fitting the sampling points through logarithmic fitting.

As illustrated by the figure, our \F is obviously faster than WDR(LSTM) when the sequence length is large than 180. LSTM-based model is fast in short sequences, and its consuming time increases rapidly as the sequence becomes longer. In actual car rides data, long-range sequences are common, so \F is more applicable for practical problems.

\section{Conclusion and Future Work}
\label{Conclusion}
In this paper, to our best knowledge, we are the first to estimate travel time entirely based on FFN with attention by presenting \F. This idea is novel and quite different from RNN based methods which have been state-of-the-art since 2018. Furthermore, we propose a novel Multi-factor self-attention mechanism for FFN to better mine sequence semantic information for ETA which is affected by various factors. 
Through sufficient experiments on the massive real-world dataset from a famous intelligent travel platform, we conclude that \F achieves slight improvements over other state-of-the-art methods regarding to the prediction precision. Most importantly, our method significantly speeds up the inference process compared with RNN based methods.
Multi-factor self-attention mechanism is also verified by experiments to be superior to the popular Multi-head self-attention that is proposed for natural language processing. Future efforts will be made to adopt our Multi-head self-attention for other sequence learning tasks which are also affected by many complex factors. Besides, we plan to conduct a series of online tests for \F and decide if we could adopt this promising deep learning framework for large scale practical application.

\section*{Broader Impact}

We present the statement of the broader impact of our paper as followed:

a) This research is benefitial for many other tasks in ITS, such as route planning, navigation and vehicle dispatching. Our \F which is an significantly faster and more accurate framework for ETA do good to the ride-hailing platforms, such as DiDi and Uber, for providing better user experiencs. Furthermore, our method promotes the long-term development of ITS and the spatial-temporal sequential prediction;

b) We are convinced that nobody will be put at disadvantage from our work. On the contrary, our research indirectly makes it more convenient for many people to travel and helps environmental protection;

c) Our framework is a potential framework for online application, which reflects the practical application value of our model. If our model is lucky enough to be selected as the practical application method, the ride-hailing platform will also go through multi-directional tests in order to avoid the only economic loss once our method fails;

d) We ensure that the method does not leverage any biases in the data. The experiments are carried out on the large-scale real-world vehicle travel dataset. The data includes more than 500 million trajectories and covers almost all road types. Therefore, the distribution tends to be that of real world.

\small
\bibliographystyle{plain}
\bibliography{neurips_2020}







\end{document}